# A Novel Approach to Texture Classification using Statistical Feature

B. Vijayalakshmi[1], V. Subbiah Bharathi[2]

[1]Asst. Prof., Dept. of MCA, Velammal College of Management and Computer Studies
vlakshmi752002@yahoo.co.in
[2]Principal, DMI College of Engineering
yughasurya@yahoo.co.in

**ABSTRACT:-**

Texture is an important spatial feature which plays a vital role in content based image retrieval. The enormous growth of the internet and the wide use of digital data have increased the need for both efficient image database creation and retrieval procedure. This paper describes a new approach for texture classification by combining statistical texture features of Local Binary Pattern and Texture spectrum. Since most significant information of a texture often appears in the high frequency channels, the features are extracted by the computation of LBP and Texture Spectrum and Legendre Moments. Euclidean distance is used for similarity measurement. The experimental result shows that 97.77% classification accuracy is obtained by the proposed method.

## 1. INTRODUCTION

Texture analysis is one of the most fundamental tasks used in content based image retrieval, which specifies the characteristics of image content without any additional caption or text information. Features may be text based features or visual features. The color, texture and shape features are general features and others are domain specific features. Feature extraction is the basis for Content based image retrieval.

Texture contains important information about the structural arrangement of surfaces and their relationship to the surrounding environment. A variety of techniques ranging from statistical methods to multi resolution filtering have been developed for texture analysis. In texture analysis, the most important task is to extract texture features, which specifies textural characteristics of the original image. Tuceryan and Jain [22] divided texture analysis methods into statistical, geometrical, model based and signal processing. Recent study of human vision system indicates that spatial/frequency representation which preserves both global and local information is adequate for quasi periodic signal. This observation has motivated researchers to develop multi resolution texture models.

In the early 70's Haralick et al [5] proposed coocurrence matrix representation of texture feature. This approach explored gray level spatial dependent of texture. Tamura et al [2] explored texture representation from different angle and proposed a computational approximation on six visual properties like coarseness, contrast, directionality, linelikeness, regularity and roughness. The QBIC system and MARs system further improved Tamura's texture representation. In the early

90's, the wavelet transform was introduced for texture representation. Smith and Chang [28, 29] used the statistics such as mean and variance features are extracted from wavelet subbands as texture representation. Gross et al [4] used Wavelet Transform together with KL expansion and kohenon maps to perform texture analysis. Thygagarajan et al [30] and Kundu et al combined wavelet transform with coocurrence matrix to take the advantages of statistics based and transform based texture analysis. Ma and Manjunath [8] evaluated texture image annotation by using various wavelet texture representation including orthogonal and bi-orthogonal wavelet transform, tree structured wavelet transform and Gabor wavelet transform.

The texture spectrum was initially used as a texture filtering approach (He and Wang, 1991). The importance of the texture spectrum method is determined by the extraction of local texture information for each pixel and of the characterization of textural aspect of a digital image in the form of a spectrum. Also, Ojala et al. [30] proposed the uniformed local binary patterns (LBP) approach to extracting rotation and histogram equalization invariant features, which was extended by Huang, Li and Wang by computing the derivative-based local binary patterns and applied it to the application of face alignment. The approach of the conventional LBP is simple and efficient which considers the uniform patterns in the images which will be local features of an image.

Moment functions of the two dimensional image intensity distribution are used in a variety of applications like visual pattern recognition, object classification, template matching, edge detection, pose estimation, robotic vision and data compression. Image moments that are invariant with respect to the transformations of scale, translation and rotation find applications in areas such as pattern recognition, object identification and template matching. Orthogonal moments have additional properties of being robust in the presence of image noise and have a near zero redundancy measure in a feature set.

The statistical methods and moment based feature extraction is proposed in this paper. The objective of this paper is to retrieve images accurately with the combined texture features. The texture features of Local Binary pattern and Texture spectrum is combined and it provides better texture classification. The classes of texture features are investigated in classification experiments with arbitrary texture images taken from Brodatz album [12]. The classification accuracy rates are compared with that of Local Binary Pattern, Legendre Moment and Texture Spectrum methods and the results are found to be improved significantly.

This paper is organized as follows: Second section explains the model behind our work and the method of detecting the presence of texture features. In the third section, with the proposed set of textured images are represented by the global descriptor, namely Texture Spectrum, Legendre Moment and Local Binary Pattern and the experimentation with a set of standard Brodatz Textural Album. Finally, the conclusion about our approach and its application for unsupervised texture classification has been highlighted along with further scope of this work.

## 2. METHODOLOGY

**2.1 Proposed Framework:-**

The process involved in the methodology is depicted in the Figure1. Initially each input images of $512 \times 512$ texture images are normalized to $64 \times 64$ of 64 sub images. Local Binary pattern is applied to each sub images and the resultant images are stored into the database. Randomly selected 10 sub images are considered as training set and remaining sub images are testing set. Compute the feature vectors for training set of images by applying Texture spectrum approach to calculate the Texture Unit $(TU)$ number and then compute histogram and calculate mean of a feature vector. This process is repeated to compute the feature vector for testing set of images. Euclidean distance is used to determine the similarity measure between the trained and tested image set. Finally images are classified based on the Minimum Distance Decision rule.

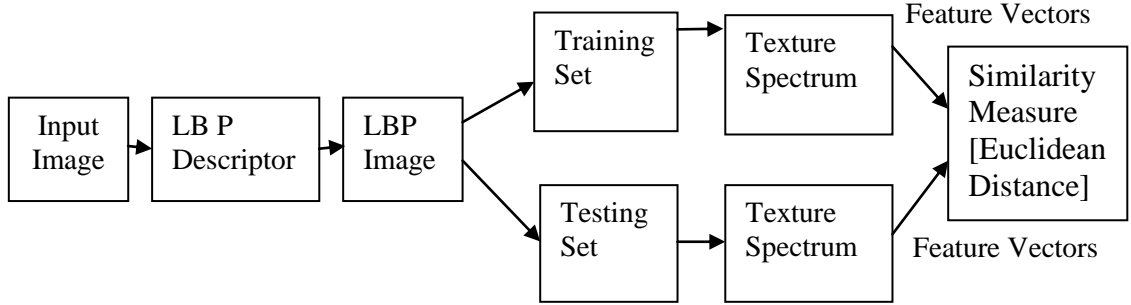

Figure 1. Proposed Architecture

## 2.2 Local Binary Pattern

Local Binary Pattern operator is a simple yet very efficient texture operator which labels the pixels of an image by thresholding the neighborhood of each pixel with the value of the center pixel and considers the result as a binary number. The original LBP method is a complementary measure for local image contrast. LBP extracts texture feature in spatial domain. The LBP value is determined as:

$$\text{LBP} = \sum_{i=1}^{8} E_i . 2^{i-1} \qquad (1)$$

where

$$E_i = \begin{cases} 1 & \text{if } V_i \geq V_0 \\ 0 & \text{if } V_i < V_0 \end{cases} \qquad (2)$$

For each input image which is shown in figure 2(a), local binary pattern is applied and the transformed images are shown in figure 2(b).

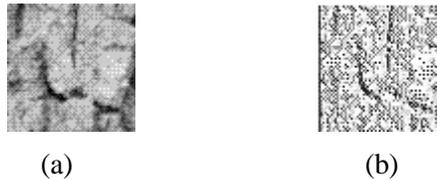

(a)　　　　　　　　(b)
Figure 2: Transformation of Input image (a) into LBP image (b)

## 2.3 Texture Spectrum

In a square raster digital image, each pixel is surrounded by 8 neighboring pixels. The local texture information for a pixel can be extracted from a neighborhood of $3 \times 3$ pixels, which represents the smallest complete unit. We define the corresponding texture unit by a set containing 8 elements such as: $TU = \{E_1, E_2, ..... E_8\}$ where $E(i = 1,2,.....8)$ is determined by the formula:

$$E_i = \begin{cases} 0 \text{ if } V_i < V_0 \\ 1 \text{ if } V_i = V_0 \\ 2 \text{ if } V_i > V_0 \end{cases} \quad (3)$$

For $\{i = 1, 2... 8\}$ and the element $E_i$, occupies the same position as the pixel i. As each element of TU has one of three possible values, the combination of all eight elements results in $3^8 = 6561$ possible texture units in total. The texture unit number is calculated by the following formula:

$$N_{TU} = \sum_{i=1}^{8} E_i . 3^{i-1} \quad (4)$$

Thus the statistics on frequency of occurrence of all the texture units over a large region of an image should reveal texture information. For each input image which is shown in Figure 3(a), texture spectrum methodology is applied and the transformed images are shown in Figure 3(b).

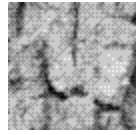 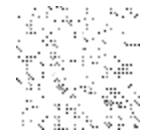

(a)                      (b)

Figure 3: Transformation of Input image (a) into Texture spectrum image (b)

### 2.4 Legendre moment

The moments with Legendre polynomials as kernel functions denoted as Legendre moments were introduced by Teague [21]. Legendre moments belong to the class of orthogonal moments and they were used in several pattern recognition applications. They can be used to attain a near zero value of redundancy measure in a set of moment functions so that the moments correspond to independent characteristics of the image. The definition of Legendre moments has a form of projection of the image intensity function into Legendre polynomials. The two-dimensional Legendre moments of order (p + q), with image intensity function f(x, y), are defined as

$$L_{pq} = \frac{(2p+1)(2q+1)}{4} \int_{-1}^{1}\int_{-1}^{1} P_p(x)P_q(y) f(x, y) dx dy \quad (5)$$

where x, y ε {-1, 1}. Since the region of definition of Legendre polynomials is the interior of {-1, 1}, a square image of N x N pixels with intensity function f(i, j), $0 \le i, j \le (N-1)$ is scaled in the region of -1 < x, y < 1 as a result of this, equation (5) can now be expressed in discrete form as

$$L_{pq} = \lambda_{pq} \sum_{i=0}^{N-1} \sum_{j=0}^{N-1} P_p(x_i) f(i,j) \qquad (6)$$

where the normalizing constant,

$$\lambda_{pq} = \frac{(2p+1)(2q+1)}{N^2} \qquad (7)$$

$x_i$ and $y_i$ denote the normalized pixel coordinated in the range of {-1, 1} which are given by

$$x_i = \frac{2i}{N-1} - 1 \quad \text{and} \quad y_i = \frac{2i}{N-1} - 1 \qquad (8)$$

The Legendre polynomial of order 10 is applied to the input images and its feature vectors are computed for further classification.

## 3. RESULT AND DISCUSSION

In order to evaluate the texture features using Local Binary Pattern and Texture Spectrum for texture characterization and classification, several experimental studies carried out on 12 texture classes of brodatz texture images of $512 \times 512$ pixels which are depicted in Figure 4. These texture images can be normalized to $64 \times 64$ pixels with 768 samples of sub images. Classification experiments were performed by using texture features extracted using Local Binary Pattern (LBP) histogram descriptors, Legendre Moment (LM) and Texture Spectrum (TS) histogram descriptors separately. The histogram of LBP and Texture spectrum descriptor is shown in Figure 5. Then classification performance was analyzed by extracting texture spectrum on LBP images (LBPTS), Legendre Moment on LBP images (LBPLM) and Texture spectrum descriptor with Legendre Moment (TSLM).

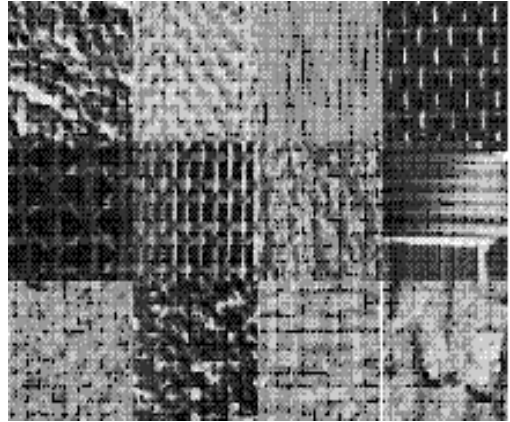

Figure 4: Brodatz Images for Texture Classification

A quantitative study was performed using classification over 12 texture images based on the proposed method. The Euclidean Distance is used to determine the distance between training and testing set. The minimum distance decision rule is used for the texture classification. The classification accuracy of texture features of the various methods are shown in the Table1 and its performance is depicted as graph in Figure 6. The experimental result shows that the combined features of Local Binary pattern and Texture spectrum descriptor approach has better classification performance than Local Binary pattern, Legendre Moment and Texture Spectrum.

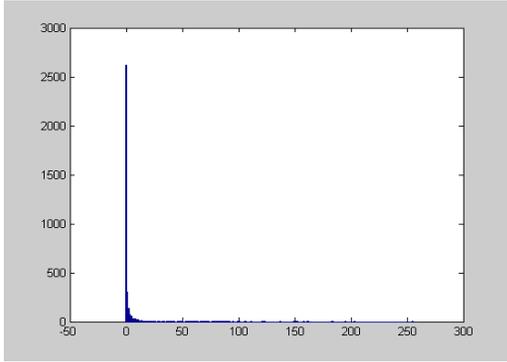 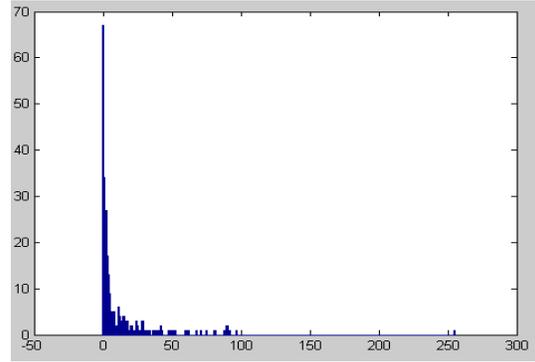

a) Histogram of Texture spectrum        b) Histogram of Local Binary Pattern

Figure 5: Histogram for Bark texture using Texture spectrum and Local Binary Pattern descriptor

Table1: Classification Accuracy (%) of the Texture Features

| Texture Classes | LM | LBP | TS | TSLM | LBPLM | LBPTS |
|---|---|---|---|---|---|---|
| D6 | 100 | 96.3 | 98.15 | 11 | 42.6 | 87.03 |
| Handpaper | 100 | 100 | 98.15 | 30 | 74.1 | 98.15 |
| Sand | 92.59 | 96.3 | 100 | 37 | 72.22 | 100 |
| Water | 88.88 | 100 | 68.52 | 38.88 | 98.15 | 94.44 |
| Bark | 64.81 | 92.6 | 98.15 | 38.88 | 61.1 | 100 |
| Leather | 87.04 | 100 | 100 | 53.7 | 59.3 | 100 |
| Weave | 96.29 | 90.7 | 100 | 62.96 | 53.7 | 100 |
| Raffia | 55.55 | 98.2 | 92.59 | 70.37 | 77.8 | 98.15 |
| PC | 81.48 | 79.6 | 100 | 83.33 | 72.22 | 100 |
| D53 | 96.29 | 100 | 100 | 100 | 98.15 | 100 |
| Average Precision | 86.293 | 95.37 | 95.556 | 52.612 | 70.934 | **97.777** |

## 4. CONCLUSION

The feature extraction for texture analysis has been performed with the Legendre Moment, Local Binary Pattern images, Texture Spectrum, Texture Spectrum with Legendre Moment, LBP with Legendre Moment and proposed method. The experimental evaluation shows that Local Binary pattern is able to extract spatial texture features and it has promising discriminating performance for different textures. The texture spectrum extracts the global features. The combined approach of LBP images with Texture Spectrum descriptor performs better classification than the individual features and it provides 97.77% classification accuracy. In the future work, various other features can be combined with the proposed method and applied for various applications such as medical image retrieval.

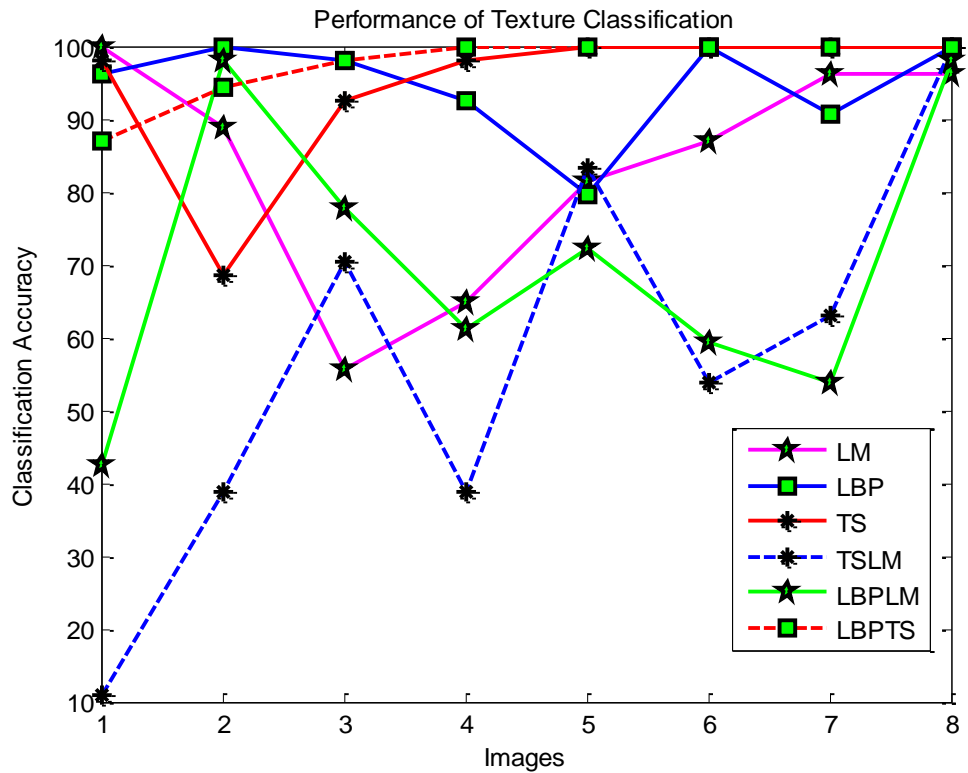

Figure 6: Graph representing the performance analysis of Texture Feature

Authors

B. Vijayalakshmi is currently working as Assistant Professor in MCA Department of Velammal College of Management and Computer Studies, Surapet, Tamil Nadu, India. She is pursuing Ph.D in Mother Theresa Women's University, Kodaikanal, Tamil Nadu, India. She received M. Phil degree from the same University and received M.C.A from Madras University.

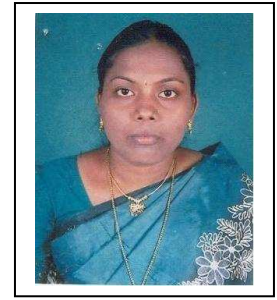